\documentclass[11pt]{article}

\usepackage[letterpaper,margin=1in]{geometry}
\usepackage[T1]{fontenc}
\usepackage[utf8]{inputenc}
\usepackage{times,helvet,courier}
\usepackage[hyphens]{url}
\usepackage{graphicx,booktabs,amsmath,amssymb,amsfonts,xspace,multirow}
\usepackage{subcaption}
\usepackage{authblk}
\usepackage{microtype}
\usepackage[hidelinks]{hyperref}
\usepackage{marvosym}

\newcommand{\method}{SED-MCTS\xspace}
\title{An Interpretable Approach to PDE Solution Discovery via Structural Experience Distillation}
\author[]{Yunpeng Gong}
\author[]{Huolong Wu}
\author[]{Can Yang}
\author[]{Min Jiang\textsuperscript{\Letter}}
\affil[]{School of Informatics, Xiamen University}
\date{}
\hypersetup{
  pdftitle={An Interpretable Approach to PDE Solution Discovery via Structural Experience Distillation},
  pdfauthor={Yunpeng Gong, Huolong Wu, Can Yang, Min Jiang}
}

\begin{document}
\maketitle

\begin{abstract}
	PDE solution discovery aims to identify explicit symbolic expressions for unknown physical fields from observations under known physical constraints. Existing methods, however, collapse data fidelity and physical consistency into a single terminal score used as the sole feedback signal, providing little information about which subexpressions are responsible for a candidate's final performance. This opaque terminal feedback severely limits the interpretability of the search process itself, offering no insight into why a candidate succeeds or fails. Consequently, reusable structures in otherwise suboptimal candidates are often discarded, whereas incidental syntax along successful search trajectories may be repeatedly reinforced. We propose SED-MCTS, a Monte Carlo tree search approach that distills structural experience from evaluated expressions and reuses it to guide subsequent symbolic solution search. Through counterfactual subtree interventions, SED-MCTS estimates local structural contributions, routes reliable evidence to the responsible construction edges, and preserves useful components in a refined structural archive. The approach naturally extends to coupled multiphysics systems. Across a diverse suite of PDE benchmarks, SED-MCTS achieves strong performance under a fixed evaluation budget and improves search efficiency and robustness under noisy or scarce observations. 
\end{abstract}

\section{Introduction}

Symbolic regression (SR) seeks explicit mathematical expressions directly
from observational data, yielding models that can be inspected, simplified,
and reused in scientific reasoning~\cite{koza1992genetic,
	schmidt2009distilling,udrescu2020aifeynman}. In conventional SR, a candidate
expression is judged primarily by how accurately its values reproduce a
finite set of input--output observations. Consequently, many structurally
different expressions may achieve similar data-fitting accuracy, especially
when observations are sparse or noisy, even though their extrapolation
behavior and scientific meaning can differ substantially.

Physics-informed symbolic regression imposes a stronger requirement. For a
system governed by partial differential equations (PDEs), the desired
expression must not only reproduce the observed field values, but also remain
consistent with the governing differential operator throughout the domain
and, where imposed, with the corresponding initial and boundary conditions.
Numerical solvers provide discrete approximations of such fields
~\cite{leveque2007finite,hughes1987finite,trefethen2000spectral}, whereas a
symbolic solution must encode the underlying spatiotemporal structure in a
compact closed-form expression. A candidate may reproduce all available
observations while still exhibiting nonzero PDE residuals at unobserved
collocation points. Physics-informed symbolic search must therefore identify
expressions that simultaneously reduce observation error and differential
residual~\cite{raissi2019physics}, while controlling expression complexity.

This additional physical requirement poses new challenges for symbolic
search, rather than merely adding another evaluation term. Candidate quality
is determined not only by the values produced by an expression, but also by
the behavior of the complete expression under the governing differential
operator. A local symbolic component may contribute differently to data
accuracy and physical consistency: it may have only a limited effect on the
observed values while being essential for reducing the PDE residual, or it
may improve data fitting while introducing physically inconsistent
derivatives. Existing methods typically aggregate data accuracy and physical
consistency into a single value, failing to reveal how individual symbolic
components contribute to either objective. This difficulty becomes more
pronounced in coupled multiphysics problems, where multiple field
expressions must be constructed jointly and a subtree in one field may
affect the residuals of other fields through the coupled equations, further
exacerbating the complexity of the problem.

The above discussion reveals a fundamental difficulty: for a complete
candidate expression, its terminal score, whether data-fitting error or PDE
residual, is provided as a single aggregated value, with no indication of
which specific substructures are responsible for the final performance.
However, symbolic regression is inherently a compositional construction
process, in which the contributions of individual local components
(subexpressions) to the overall performance are often highly uneven. A
poorly scoring expression may internally contain physically plausible and
reusable substructures; conversely, a well-scoring expression may contain
redundant or even detrimental syntactic components. Therefore, how to
identify and extract valuable local structural information from the aggregate
terminal feedback of complete expressions, referred to as the \emph{subtree
	credit assignment} problem, remains a key unresolved challenge in
physics-informed symbolic regression.

Closely related to the credit assignment problem is the effective reuse of
structural knowledge. Symbolic regression search typically maintains a
population or a set of candidate solutions, yet most existing methods retain
only a few optimal complete expressions after evaluation, discarding
candidates with lower overall ranks even if they contain high-quality
substructures internally. These discarded components may be precisely the key
building blocks for solving certain subproblems, such as temporal decay
terms, spatial oscillatory factors, or cross-coupling terms. If such useful
structures scattered across different candidates could be accumulated and
reused throughout the search, both search efficiency and solution quality
would be significantly improved. Nevertheless, existing symbolic regression
frameworks lack a systematic mechanism to convert one-shot terminal
evaluations into structural knowledge that can be persistently accumulated
and reused.

Monte Carlo tree search (MCTS) provides an attractive foundational framework
for addressing these challenges. MCTS constructs expressions by incrementally
expanding grammar-valid prefixes, and its stepwise decision-making process
naturally aligns with the compositional nature of symbolic expression
construction. More importantly, MCTS maintains a progressively expanding tree
structure during search, which offers the potential to decompose terminal
feedback into local construction decisions: each node, corresponding to a
partial expression, and each edge, corresponding to a grammar action, can be
assigned individual statistics. However, standard MCTS applied to symbolic
regression still primarily uses complete expressions through a delayed scalar
return that is backed up uniformly along the entire construction path. This
treatment fails to exploit the tree structure of MCTS to address the subtree
credit assignment problem: it cannot distinguish the actual contributions of
different construction steps to the final return, nor can it extract and
accumulate local structural knowledge from individual candidates.

To address these two interconnected problems of subtree credit assignment
and effective reuse of structural knowledge, we propose SED-MCTS, where SED
denotes structural experience distillation. Rather than using terminal
candidates only through their scalar returns, SED-MCTS treats every evaluated
expression as a potential source of reusable structural knowledge. It
estimates the contribution of local subtrees through physics-informed
counterfactual interventions, filters weak or redundant components, and
routes reliable evidence only to the prefix-action edges responsible for
generating useful structures. The accumulated responsible-edge statistics
define an endogenous search prior, while a refined elite-diverse archive
preserves high-quality components for subsequent mutation and crossover. In
this way, terminal evaluation is transformed from a one-time scalar judgment
into persistent structural experience that continuously improves subsequent
expression construction. The same framework naturally extends to coupled
multiphysics problems through its conditional attribution mechanism; detailed
implementations are deferred to Section~3.

Our contributions are threefold. First, we formulate physics-informed structural distillation as a post-terminal knowledge-reuse problem for MCTS-based symbolic regression, replacing undifferentiated trajectory-level feedback with localized structural evidence. Second, we introduce responsible-edge statistics and a refined structural archive that jointly guide forward search and structural editing: the responsible-edge statistics define an endogenous prior for MCTS selection and mutation, while the archive preserves reusable components for compatible crossover. Third, we evaluate SED-MCTS on single-field PDE symbolic-solution recovery, coupled multiphysics systems, and settings with noisy or scarce observations, demonstrating improved accuracy, search efficiency under a fixed terminal-expression budget, and robustness across diverse benchmarks. The supplementary material provides pseudocode, theoretical analyses, complete PDE definitions, hyperparameter settings, recovered expressions, and extended diagnostic results.

\begin{figure*}[t]
\centering
\includegraphics[width=0.95\textwidth]{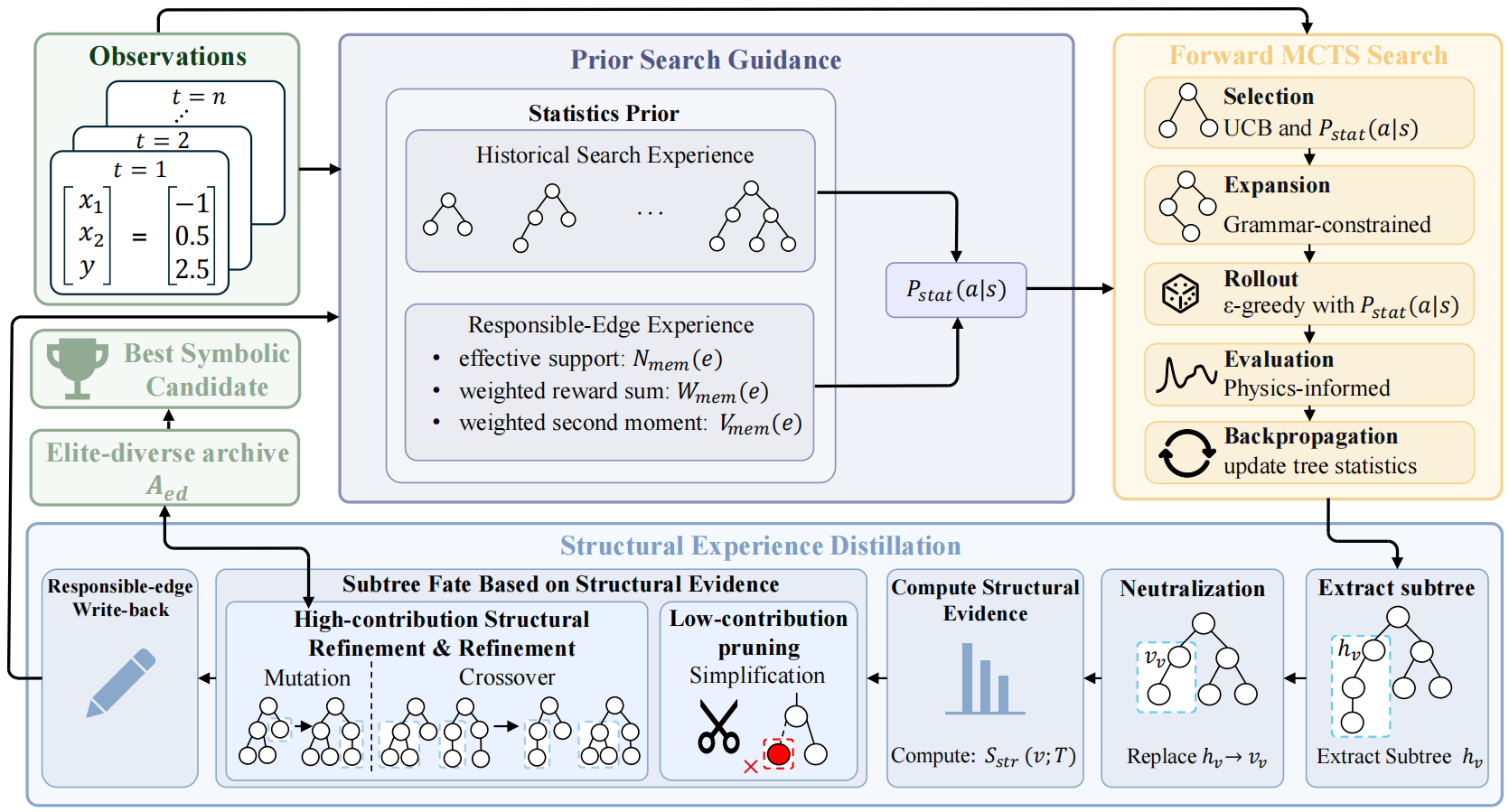}
\caption{Overview of SED-MCTS. The forward search constructs grammar-valid symbolic expressions using MCTS and evaluates each completed candidate with the joint data--physics objective. After terminal evaluation, context-preserving subtree neutralization estimates the residual contribution of editable subtrees. The resulting structural evidence is routed back to the responsible prefix--action edges to update the statistical prior, while high-contribution structures are refined through guided mutation and crossover and low-contribution structures are considered for pruning. The refined elite-diverse archive and accumulated edge statistics are then reused to guide subsequent MCTS selection, rollout, and structural editing.}
\label{fig:framework}
\end{figure*}

\section{Related Work}

Classical symbolic regression constructs expression trees through genetic
programming~\cite{koza1992genetic}, while more recent approaches improve
expression generation through risk-seeking policies, neural-guided population
initialization, and efficient evolutionary
search~\cite{petersen2021deep,mundhenk2021symbolic,cranmer2023pysr}.
Scientific symbolic discovery broadly includes two related but distinct
settings. Sparse equation-discovery methods select governing terms from a
predefined candidate library~\cite{brunton2016sindy,rudy2017pdefind},
whereas symbolic solution discovery searches directly for a closed-form
function represented by a grammar or expression tree. Physics-informed
symbolic methods further evaluate candidate expressions using
differential-equation residuals and, when applicable, initial or boundary
constraints~\cite{das2025pinnsr,majumdar2022pisn,
	mazraeh2025mctspinn,cohen2024pigp}. These approaches substantially
strengthen the terminal objective used to judge an expression, but typically
retain an aggregate candidate-level evaluation. Consequently, they provide
limited information about which internal structures are responsible for data
agreement or physical consistency.
Genetic-programming research has long exploited reusable building blocks
through subtree recombination, semantic-aware variation, component
libraries, diversity-preserving archives, and transfer across related tasks
or representations~\cite{moraglio2012geometric,krawiec2013locally,
	bruneton2019qualitydiversity,chen2022transfer,zhong2025multiform}.
Such mechanisms commonly assess component utility through the fitness of its
source individual, semantic compatibility, archive-niche quality, ancestry,
or realized transfer performance~\cite{burlacu2015effectiveness}. RSSP is
more closely related to our setting because it evaluates local
subexpressions and uses physics-informed structural sensitivity to guide
pruning and variation~\cite{gong2025rssp}. However, local sensitivity is
primarily used to modify the currently evaluated candidate, while
conventional libraries and archives preserve concrete individuals or
subtrees without explicitly converting a physics-informed terminal objective
into persistent construction-level credit. Existing work therefore offers
either local structural diagnostics without cross-candidate statistical
memory, or reusable component repositories whose contents are not grounded
in responsibility-aware physical attribution.

MCTS is well suited to symbolic regression because grammar-valid expressions
can be constructed through sequential decisions while computation is
adaptively allocated among promising
prefixes~\cite{kocsis2006bandit,kamienny2023dgsr}. Existing MCTS-based
symbolic-regression methods improve expression discovery through learned or
heuristic proposals, modified value estimates, and transitions between
promising search states~\cite{kamienny2023dgsr,huang2025improving}.
Nevertheless, their feedback is predominantly associated with complete
candidates, prefixes, or entire construction trajectories. Standard backup
consequently propagates a terminal return through all decisions on a
successful path, without distinguishing actions that create useful local
motifs from incidental syntax elsewhere in the expression. It also provides
no direct mechanism for retaining valuable components found inside
lower-ranked candidates. \method addresses the missing connection between
structural assessment and persistent reuse by converting evaluated
candidates into structure-aware search experience that guides subsequent
tree construction and compatible component-level variation. This positioning
distinguishes \method from trajectory-level MCTS feedback and from
conventional archives, while leaving the specific attribution and reuse
mechanisms to the method section.

\section{Method}
\label{sec:method}

SED-MCTS converts terminal evaluations into reusable structural experience. For each completed expression, a context-preserving neutralization operation produces node-level evidence from the joint data--physics objective. This evidence is routed to the grammar edges that generated the corresponding subtree and is also stored in a refined structural archive. The accumulated experience then biases subsequent MCTS construction and lightweight structural editing. We use \emph{structural evidence} for attribution extracted from one terminal expression and \emph{structural experience} for evidence accumulated across expressions.

\subsection{Physics-Informed Symbolic Search}
\label{sec:search}

Let $z=(\mathbf{x},t)$ denote the spatiotemporal coordinates and $\mathcal D_d=\{(z_i,y_i)\}_{i=1}^{n_d}$ the observations. Given a differential operator $\mathcal N$, we search a typed grammar for an expression $f_\theta$, where $\theta$ collects all free constants. With residual points $\{\tilde z_j\}_{j=1}^{n_r}$, we define
\begin{align}
	D_{\rm data}(f)&=\frac{1}{n_d}\sum_{i=1}^{n_d}\frac{|f(z_i)-y_i|^2}{\sigma_y^2+\epsilon},\\
	D_{\rm phys}(f)&=\frac{1}{n_r}\sum_{j=1}^{n_r}\frac{|\mathcal N[f](\tilde z_j)|^2}{\sigma_r^2+\epsilon},\\
	L(f)&=D_{\rm data}(f)+\lambda_{\rm pde}D_{\rm phys}(f).
	\label{eq:terminal_loss}
\end{align}
The two terms are separately variance-normalized, so we fix $\lambda_{\rm pde}=1$ for all tasks. We use the same numerical stabilizer $\epsilon=10^{-8}$ throughout. After optimizing $\theta$, the terminal reward is $R(f)=\exp[-L(f)-\eta\mathcal C(f)]$, where $\mathcal C(f)=|T_f|$ is the number of nodes in the expression tree and $\eta=10^{-2}$ is fixed across all experiments.

MCTS represents expression construction as grammar-constrained decisions. A state $s$ is a partial expression and $a\in\mathcal A(s)$ is a valid production. Let $N(s,a)$ be the visit count of edge $(s,a)$ and $N(s)=\sum_{a'\in\mathcal A(s)}N(s,a')$ the total visit count of state $s$, obtained by summing its outgoing-edge visits; $a'$ is only the summation index over valid actions. Let $M_K(s,a)$ denote the extreme-value statistic computed from the $K$ best terminal descendants of edge $(s,a)$. Selection maximizes
\begin{align}
	I(s,a)=M_K(s,a)
	&+c\left(\frac{\log(1+N(s))}{1+N(s,a)}\right)^\gamma \notag\\
	&+\lambda_s\log\!\left(P_{\rm stat}(a\mid s)+\epsilon\right),
	\label{eq:mcts_selection}
\end{align}
where $c>0$ and $\gamma\in(0,1]$ control exploration, $\lambda_s\ge 0$ weights the structural prior, and $\epsilon$ prevents undefined logarithms and divisions. The state-conditioned prior $P_{\rm stat}$ is defined below. Each completed expression is evaluated by Eq.~\eqref{eq:terminal_loss} and then structurally distilled.

Standard MCTS backs up one terminal reward along the whole trajectory and therefore cannot identify which local decisions created useful structure. Motivated by numerically stable masking strategies~\cite{gong2025rssp}, we instead neutralize every editable subtree and measure the resulting data--physics response.

Let terminal tree $T$ represent $f_T=C_v[h_v]$, where $h_v$ is the subtree rooted at node $v$ and $C_v[\cdot]$ its surrounding context. We define the neutralized expression
\begin{equation}
	\bar f_v := C_v[\nu_v],
	\label{eq:neutralized_expression}
\end{equation}
where $\nu_v$ removes the subtree's input-dependent response while preserving type validity and numerical scale. For a binary-rooted subtree $h_v=\psi(h_l,h_r)$, we use $\nu_v=0$ for $\psi\in\{+,-\}$ and $\nu_v=1$ for $\psi\in\{\times,\div\}$. For a unary subtree $h_v(z)=\phi(g(z))$, let $\mathcal Z=\{z_i\}_{i=1}^{n_d}\cup\{\tilde z_j\}_{j=1}^{n_r}$ and
\begin{equation}
	\bar g_v=\frac{1}{|\mathcal Z|}\sum_{z\in\mathcal Z}g(z),\qquad
	\nu_v=\phi\!\left(\Pi_{\Omega_\phi}(\bar g_v)\right),
	\label{eq:unary_neutralization}
\end{equation}
where $\Pi_{\Omega_\phi}$ projects the reference input to the numerically safe domain of $\phi$. Variable leaves are replaced by their empirical mean over $\mathcal Z$, while pure constants are not attributed separately. All fitted constants remain frozen during neutralization.

The response of node $v$ is evaluated through the two objective channels,
\begin{align}
	\mathbf r(v;T)&=
	\begin{bmatrix}
		r_{\rm data}(v;T)\\
		r_{\rm phys}(v;T)
	\end{bmatrix}
	=
	\begin{bmatrix}
		D_{\rm data}(\bar f_v)-D_{\rm data}(f_T)\\
		D_{\rm phys}(\bar f_v)-D_{\rm phys}(f_T)
	\end{bmatrix},\\
	S_{\rm str}(v;T)&=\frac{[r_{\rm data}(v;T)+r_{\rm phys}(v;T)]_+}{(1+|h_v|)^\zeta},
	\label{eq:structural_evidence}
\end{align}
where $[x]_+=\max(x,0)$ and $|h_v|$ is the subtree size. A positive score means that neutralizing $h_v$ worsens the complete objective. We evaluate every non-root node for which a valid neutralization is defined; their set is denoted by $V(T)$. To make trees of different sizes comparable, the evidence is normalized within each terminal expression:
\begin{equation}
	\rho_v(T)=\frac{S_{\rm str}(v;T)}{\sum_{v'\in V(T)}S_{\rm str}(v';T)+\epsilon}.
	\label{eq:within_tree_evidence}
\end{equation}
Here $v$ is the currently evaluated subtree root and $v'$ indexes all nodes in $V(T)$ for summation in the denominator.

For coupled fields $\mathbf f=(f_1,\ldots,f_K)$, only the selected subtree in field $f_k$ is neutralized; all other fields and fitted constants are fixed, so the resulting response includes both within-field and cross-field effects.

\subsection{Responsible-Edge Experience and Statistical Prior}
\label{sec:edge_memory}

For each node $w$ in $T$, let $e_T(w)=(s_w,a_w)$ be the grammar edge that instantiated it. If $\mathcal V(h_v)$ is the node set of subtree $h_v$, its responsible-edge set is
\begin{equation}
	\Gamma_T(v)=\{e_T(w):w\in\mathcal V(h_v)\}.
	\label{eq:responsible_edges}
\end{equation}
Thus, node-level evidence is assigned only to decisions that actually generated the attributed subtree. With $R_T := R(f_T)$, the share of node $v$ assigned to an edge $e$ is
\begin{equation}
	w_{v,e}(T)=\frac{\rho_v(T)}{|\Gamma_T(v)|}\,\mathbb I[e\in\Gamma_T(v)],
	\label{eq:edge_weight}
\end{equation}
where $\mathbb I[\cdot]$ is the indicator function. For every grammar edge $e=(s,a)$, we maintain effective support $N_{\rm mem}(e)$, weighted reward sum $W_{\rm mem}(e)$, and weighted second moment $V_{\rm mem}(e)$, updated after terminal tree $T$ by
\begin{equation}
	\begin{bmatrix}
		N_{\rm mem}(e)\\
		W_{\rm mem}(e)\\
		V_{\rm mem}(e)
	\end{bmatrix}
	\leftarrow
	\begin{bmatrix}
		N_{\rm mem}(e)\\
		W_{\rm mem}(e)\\
		V_{\rm mem}(e)
	\end{bmatrix}
	+
	\sum_{v\in V(T)}w_{v,e}(T)
	\begin{bmatrix}
		1\\
		R_T\\
		R_T^2
	\end{bmatrix}.
	\label{eq:memory_update}
\end{equation}
The corresponding mean, variance, and lower-confidence score are
\begin{align}
	\mu_{\rm mem}(e)&=\frac{W_{\rm mem}(e)}{N_{\rm mem}(e)+\epsilon},\\
	\sigma_{\rm mem}^2(e)&=\max\!\left\{0,\frac{V_{\rm mem}(e)}{N_{\rm mem}(e)+\epsilon}-\mu_{\rm mem}^2(e)\right\},\\
	q_{\rm stat}(e)&=\mu_{\rm mem}(e)-\frac{\sigma_{\rm mem}(e)}{\sqrt{N_{\rm mem}(e)+1}}.
	\label{eq:memory_statistics}
\end{align}
The second term, $\sigma_{\rm mem}(e)/\sqrt{N_{\rm mem}(e)+1}$, is the uncertainty penalty: it lowers the score of edges supported by few or highly variable observations. No additional tunable coefficient is introduced. For $a\in\mathcal A(s)$, the state-conditioned prior is
\begin{equation}
	P_{\rm stat}(a\mid s)=0.95\frac{\exp(q_{\rm stat}(s,a))}{\sum_{a'\in\mathcal A(s)}\exp(q_{\rm stat}(s,a'))}+\frac{0.05}{|\mathcal A(s)|}.
	\label{eq:statistical_prior}
\end{equation}
We fix the softmax temperature to one and the uniform component to $0.05$ in all experiments. Unseen edges have zero statistics and therefore receive a uniform prior.

\subsection{Refined Structural Archive}
\label{sec:archive}

The edge memory records how useful subtrees are generated, whereas the archive stores concrete subtrees for reuse. For each positively attributed subtree, we store its canonical form, return type, active variables, depth, complexity, source reward $R_T$, normalized evidence $\rho_v(T)$, and responsible-edge signature. Fitted constants are metadata only; commutative children are canonically ordered, and grammar-equivalent or constant-only variants are merged. The archive is partitioned by return type and complexity; within each partition, components with large $R_T\rho_v(T)$ are retained while structurally novel operator and variable patterns are preserved. Removing an archive entry does not remove its grammar actions or shrink the search space.

\subsection{Prior-Guided Structural Editing}
\label{sec:editing}

Accumulated experience is reused through bounded simplification, mutation, and crossover. High-attribution subtrees are useful but not necessarily locally optimal; we therefore preserve their parent context, return type, and principal variable dependencies as a semantic anchor while generating diverse local variants. This protected diversification avoids repeatedly reusing one motif and helps the search escape local optima. Low-attribution regions are considered for type-valid lower-complexity replacements, whereas regions with sparse or conflicting evidence remain editable to prevent early memory from becoming self-reinforcing.

Mutation regenerates a local subtree production by production using $P_{\rm stat}$, including its fixed uniform exploration component. Crossover inserts a concrete donor subtree from the refined archive after checking return type, available variables, grammar role, depth, and complexity. Thus, mutation reuses statistical knowledge of how useful structures are constructed, while crossover reuses previously validated components. Every edited expression is reoptimized and reevaluated by the complete objective in Eq.~\eqref{eq:terminal_loss}; simplification is accepted only when complexity decreases with negligible loss increase, and mutation or crossover only when the complexity-regularized reward improves. The realized utility is assigned to newly instantiated recipient edges, while the donor's historical evidence is retained. Detailed operators, budgets, archive capacity, and acceptance thresholds are given in the supplement.

\begin{figure*}[t]
	\centering
	\begin{minipage}{0.48\textwidth}
		\centering
		\includegraphics[width=\linewidth]{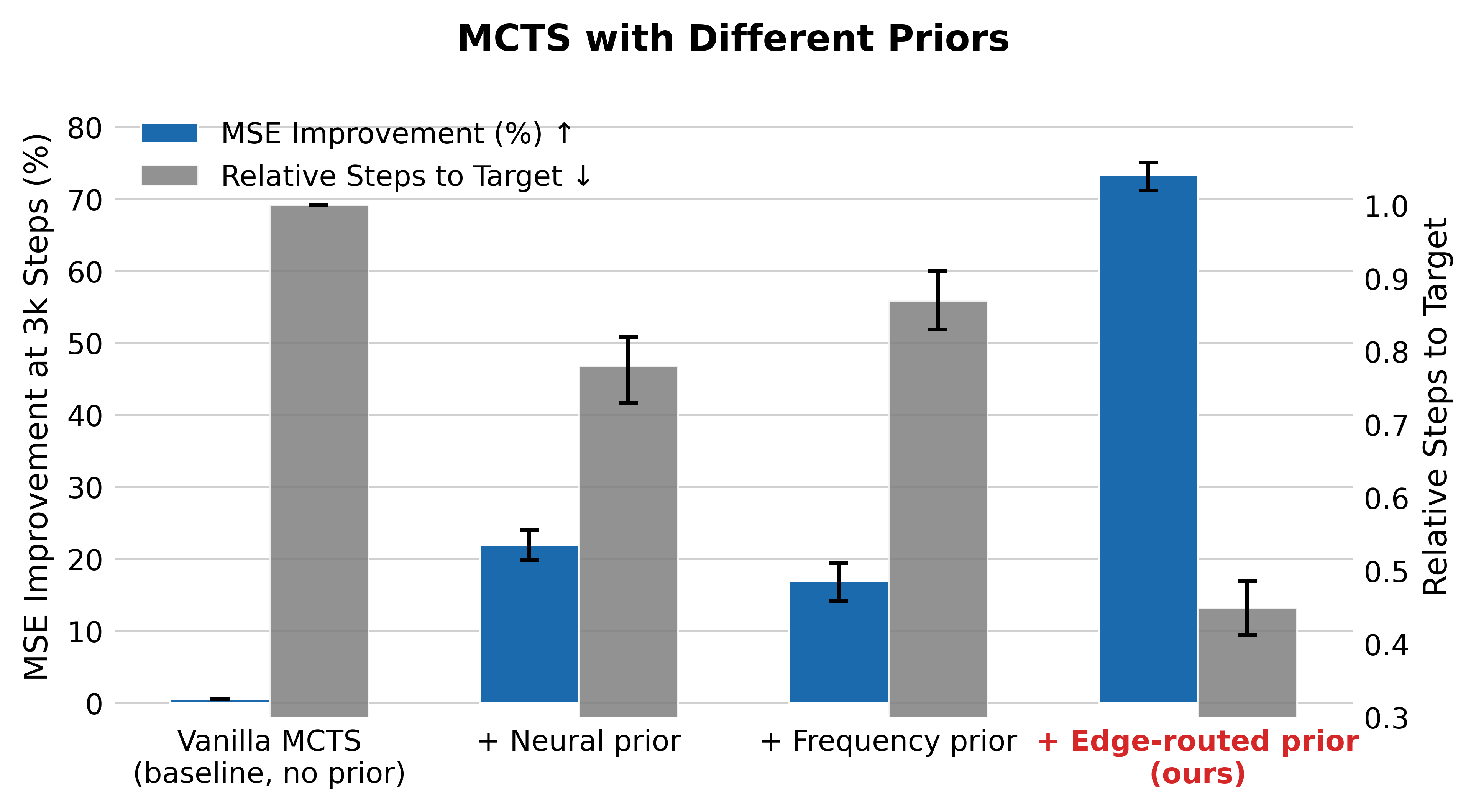}
		\subcaption{Comparison of MCTS Prior Sources.}
		\label{fig:ablation}
	\end{minipage}
	\hfill
	\begin{minipage}{0.48\textwidth}
		\centering
		\includegraphics[width=\linewidth]{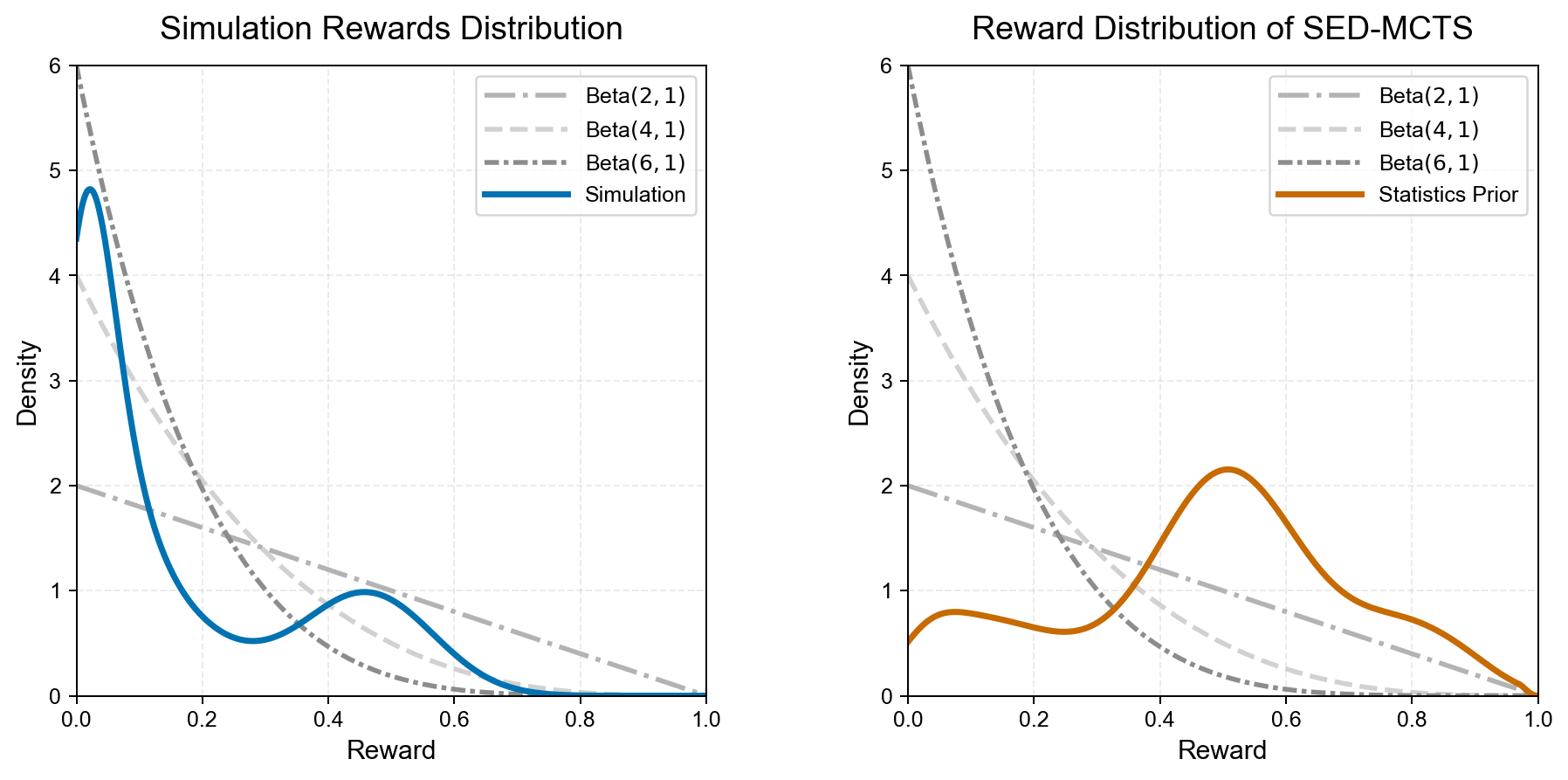}
		\subcaption{Reward Distribution Comparison.}
		\label{fig:rewards}
	\end{minipage}
	\caption{Ablation study and reward analysis on the Advection equation. (a) Prior source comparison. (b) Reward density shift from Vanilla MCTS to SED-MCTS.}
	\label{fig:ablation_rewards}
\end{figure*}

\begin{figure*}[t]
	\centering
	\begin{minipage}{0.48\textwidth}
		\centering
		\includegraphics[width=\linewidth]{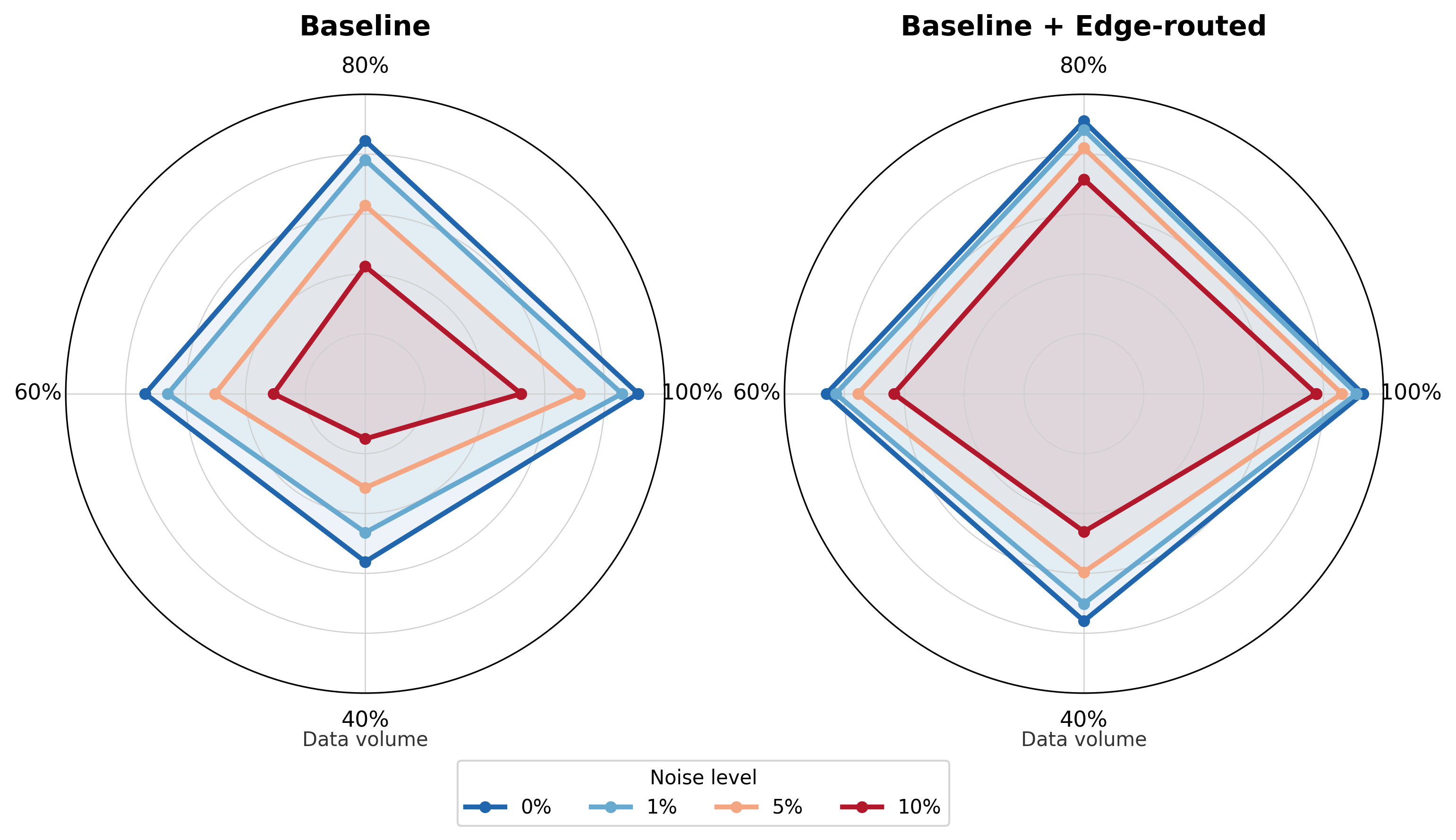}
		\caption{Robustness to Noise and Data Scarcity.}
		\label{fig:robust}
	\end{minipage}
	\hfill
	\begin{minipage}{0.48\textwidth}
		\centering
		\includegraphics[width=\linewidth]{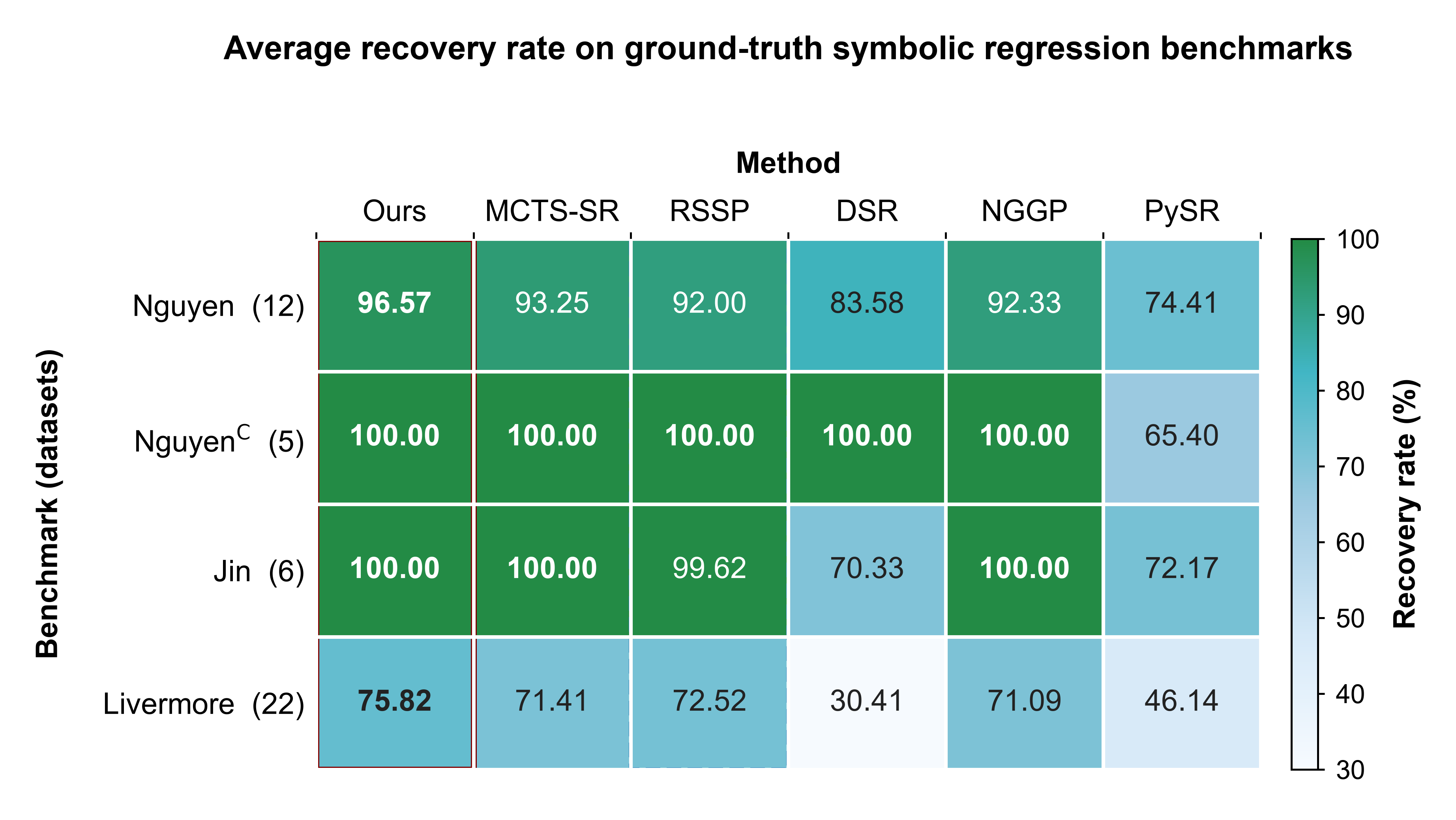}
		\caption{Exact-recovery rate on symbolic-regression suites.}
		\label{fig:gt-sr}
	\end{minipage}
\end{figure*}

\begin{table*}[t]
	\centering
	\small
	\caption{MSE on PDE benchmarks (mean $\pm$ std over 30 independent runs).}
	\label{tab:mse}
	\setlength{\tabcolsep}{4pt}
	\begin{tabular}{lccccc}
		\hline
		\textbf{PDE}
		& \textbf{MCTS-SR}
		& \textbf{PINN-SR}
		& \textbf{PISN}
		& \textbf{RSSP}
		& \textbf{Ours} \\
		\hline
		
		Advection  
		& $5.7\text{e-}10 \pm 2.3\text{e-}10$ 
		& $4.2\text{e-}10 \pm 1.8\text{e-}10$ 
		& $7.1\text{e-}11 \pm 3.5\text{e-}11$ 
		& $9.5\text{e-}14 \pm 4.1\text{e-}14$ 
		& $\mathbf{8.1\text{e-}14 \pm 3.2\text{e-}14}$ \\
		
		Diffusion  
		& $3.2\text{e-}3 \pm 1.4\text{e-}3$ 
		& $8.9\text{e-}3 \pm 4.2\text{e-}3$ 
		& $4.6\text{e-}3 \pm 2.1\text{e-}3$ 
		& $5.7\text{e-}4 \pm 2.5\text{e-}4$ 
		& $\mathbf{4.3\text{e-}4 \pm 1.8\text{e-}4}$ \\
		
		Poisson2D  
		& $6.4\text{e-}3 \pm 2.9\text{e-}3$ 
		& $2.2\text{e-}3 \pm 1.1\text{e-}3$ 
		& $5.8\text{e-}5 \pm 2.6\text{e-}5$ 
		& $4.9\text{e-}6 \pm 2.2\text{e-}6$ 
		& $\mathbf{2.8\text{e-}6 \pm 1.3\text{e-}6}$ \\
		
		Poisson3D  
		& $1.2\text{e-}2 \pm 5.6\text{e-}3$ 
		& $4.4\text{e-}3 \pm 2.0\text{e-}3$ 
		& $9.7\text{e-}5 \pm 4.3\text{e-}5$ 
		& $\mathbf{7.3\text{e-}6 \pm 3.1\text{e-}6}$ 
		& $8.1\text{e-}6 \pm 3.5\text{e-}6$ \\
		
		Wave2D     
		& $2.6\text{e-}1 \pm 1.1\text{e-}1$ 
		& $7.3\text{e-}2 \pm 3.4\text{e-}2$ 
		& $2.1\text{e-}1 \pm 9.2\text{e-}2$ 
		& $7.8\text{e-}2 \pm 3.6\text{e-}2$ 
		& $\mathbf{3.4\text{e-}3 \pm 1.5\text{e-}3}$ \\
		
		Wave3D     
		& $3.4\text{e-}1 \pm 1.5\text{e-}1$ 
		& $8.2\text{e-}2 \pm 3.9\text{e-}2$ 
		& $7.0\text{e-}2 \pm 3.2\text{e-}2$ 
		& $6.3\text{e-}2 \pm 2.8\text{e-}2$ 
		& $\mathbf{4.9\text{e-}3 \pm 2.1\text{e-}3}$ \\
		
		\hline
	\end{tabular}
\end{table*}

\begin{table*}[t]
	\centering
	\small
	\caption{Comparison of different algorithms on coupled PDE benchmark systems. $f_1$ and $f_2$ denote the two unknown functions in each system.}
	\renewcommand{\arraystretch}{1}
	\setlength{\tabcolsep}{5pt}
	\begin{tabular}{l c c c c c c c c}
		\toprule
		\textbf{Name} &
		\multicolumn{2}{c}{\textbf{T-NNGP}} &
		\multicolumn{2}{c}{\textbf{DeepM\&Mnet}} &
		\multicolumn{2}{c}{\textbf{PhySO}} &
		\multicolumn{2}{c}{\textbf{Ours}} \\
		& $f_1$ & $f_2$ & $f_1$ & $f_2$ & $f_1$ & $f_2$ & $f_1$ & $f_2$ \\
		\midrule
		Diffusion-convection &
		8.07e-02 & 1.44e-02 &
		{1.86e-02} & 1.52e-02 &
		\textbf{1.84e-02} & 1.41e-02 &
		2.05e-02 & \textbf{1.25e-02} \\
		
		Joule heating 1 &
		2.65e-05 & 3.16e+01 &
		4.28e-03 & 2.94e+00 &
		2.08e-05 & 7.45e+00 &
		\textbf{1.79e-05} & \textbf{8.10e-01} \\
		
		Joule heating 2 &
		2.71e-05 & 1.49e+01 &
		4.69e-03 & \textbf{2.12e-01} &
		{1.83e-05} & 1.50e+00 &
		\textbf{1.63e-05} & 7.93e-01 \\
		
		Joule heating 3 &
		1.97e-05 & 1.59e+01 &
		5.70e-03 & 2.52e+00 &
		{1.81e-05} & {5.59e+00} &
		\textbf{1.77e-05} & \textbf{1.75e-01} \\
		
		Joule heating 4 &
		3.77e-05 & 1.19e+01 &
		1.39e-02 & 3.26e+01 &
		{2.04e-05} & 6.52e+00 &
		\textbf{1.81e-05} & \textbf{1.59e-01} \\
		
		Joule heating 5 &
		\textbf{1.60e-05} & 2.41e+01 &
		7.58e-04 & \textbf{3.42e-01} &
		1.65e-05 & 7.32e+00 &
		1.63e-05 & 5.26e-01 \\
		
		Parabolic 1 &
		2.27e-01 & 1.66e-01 &
		\textbf{1.15e-01} & 6.37e-02 &
		1.29e-01 & 8.38e-02 &
		1.30e-01 & \textbf{5.52e-02} \\
		
		Parabolic 2 &
		2.14e-01 & 1.53e-01 &
		\textbf{8.87e-02} & \textbf{6.33e-02} &
		{9.04e-02} & 8.95e-01 &
		1.99e-01 & 8.95e-02 \\
		
		Parabolic 3 &
		6.48e-02 & 5.37e-01 &
		9.56e-02 & 3.68e-02 &
		{2.17e-02} & 1.47e-02 &
		\textbf{1.76e-02} & \textbf{1.38e-02} \\
		\bottomrule
	\end{tabular}
	\label{compare}
\end{table*}

\section{Experiments}
\subsection{Experimental Setup}
Six PDE benchmarks are evaluated, with complete equations, domains, and deterministic conditions provided in the supplementary material. The one-dimensional \textsc{Advection} benchmark models the linear transport of a scalar field under a constant velocity field, expressed as $\partial_t\phi+\partial_{x}\phi=0$ with $\phi(x,0)=\sin(x)$ on $x\in[0,1],\,t\in[0,1]$. Its analytical solution $\phi(x,t)=\sin(x-t)$ represents a traveling wave of preserved amplitude and shape, providing an ideal scenario for testing structural fidelity in symbolic recovery. The one-dimensional \textsc{Diffusion} benchmark satisfies $\partial_tu-3\partial_{x}^2u-u=0$ with exact solution $u(x,t)=e^{-2t}\sin(x)$, capturing the combined effects of diffusion and a linear reaction term. The \textsc{Poisson2D} benchmark solves $\partial_{x_1}^2u+\partial_{x_2}^2u=30x_1^2-7.8x_1+1$ on $[-1,1]^2$, with the manufactured solution $u(\mathbf{x})=2.5x_1^4-1.3x_1^3+0.5x_2^2-1.7x_2$ enforced via Dirichlet boundary conditions. The \textsc{Poisson3D} benchmark extends the elliptic operator to three dimensions as $\partial_{x_1}^2u+\partial_{x_2}^2u+\partial_{x_3}^2u=30x_1^2-7.8x_2+1$ on $[-1,1]^3$, with the manufactured solution $u(\mathbf{x})=2.5x_1^4-1.3x_2^3+0.5x_3^2$ also specified through Dirichlet boundary conditions. Both Poisson benchmarks feature polynomial source terms and manufactured solutions containing quadratic, cubic, and quartic components. The nonlinear \textsc{Wave2D} and \textsc{Wave3D} benchmarks take the general form $\partial_t^2u-\Delta u=\mathcal{G}(u,\mathbf{x},t)$. The two-dimensional equation contains nonlinear reaction terms $-u-u^3$ with exact solution $u(\mathbf{x},t)=\exp(x_1^2)\sin(x_2)e^{-0.5t}$, whereas the three-dimensional equation contains a quadratic term $u^2$ and uses $u(\mathbf{x},t)=\exp(x_1^2+x_3^2)\cos(x_2)e^{-0.5t}$. These benchmarks therefore cover transport, diffusion--reaction, elliptic polynomial structure, nonlinear wave dynamics, and compositions of exponential, trigonometric, polynomial, and temporal factors.

All search methods are compared under the same evaluation budget, with shared observations, residual collocation points, grammar, constant optimizer, and evaluation grid. It is important to emphasize that subtree interventions are intrinsic and bounded structural-diagnostic operations within SED-MCTS, rather than additional candidate-expression evaluations. They neither trigger new tree expansions nor require constant re-optimization. Therefore, we use the number of terminal expressions undergoing full constant optimization as the common search budget across all methods.
Baselines include PINN-SR, PISN, RSSP, and an MCTS-SR adaptation. Exact recovery is additionally tested on Nguyen, Nguyen$^{\rm C}$, Jin, and Livermore. Reported metrics cover dense-grid MSE, evaluations to a target error, exact recovery, and expression size. Mutation and crossover candidates count against the same terminal-evaluation budget. Complete PDE definitions, implementation details, hyperparameter settings, recovered expressions, and additional ablations are provided in the supplementary material.

Beyond single-field benchmarks, three coupled multiphysics systems with distinct coupling structures are considered. (1) Diffusion--Convection models passive scalar transport under an incompressible flow: $-\nabla\cdot\nabla u + \nabla v\cdot\nabla u = 0$ with $\nabla\cdot v = 0$, where $v$ is the velocity field and $u$ is the transported scalar. (2) Joule Heating couples electric and thermal fields via $-\nabla\cdot(\sigma\nabla V)=0$ and $\rho C\partial_t T - \nabla\cdot(k\nabla T)=\sigma|\nabla V|^2$, with electric potential $V$ and temperature $T$. (3) Parabolic describes a two-species diffusion--reaction system: $\partial_t u_1 = 0.024\,\partial_x^2 u_1 - F(u_1-u_2)$ and $\partial_t u_2 = 0.170\,\partial_x^2 u_2 + F(u_1-u_2)$, capturing bidirectional exchange between $u_1$ and $u_2$.

\subsection{Comparison of MCTS Prior Sources}
In Fig.~\ref{fig:ablation}, we compare four prior sources under a fixed budget of 10,000 fully evaluated expressions. All variants share the same grammar, constant optimizer, and evaluation protocol, without archive or structural editing; they differ only in how $P_{\rm stat}$ in \eqref{eq:statistical_prior} is constructed. Vanilla MCTS sets $\lambda_s = 0$ (no external prior) as baseline. Neural prior adopts the RSSP-style pretrained network that extracts Taylor-expansion features and computes a similarity metric to the target solution manifold. Frequency prior counts raw occurrences of each grammar edge among the elite set and normalizes the counts into $P_{\rm stat}$ without contribution weighting. Edge-routed prior (ours) applies counterfactual subtree neutralization  to estimate each subtree's marginal contribution and routes the credit exclusively to the responsible grammar edges. We report two metrics: MSE improvement over Vanilla (higher is better) and relative steps to reach a prescribed target accuracy normalized by Vanilla (lower is better). Experiments are conducted on the Advection equation, results show that any external prior outperforms the prior-free baseline. Frequency prior improves modestly because indiscriminate edge counting reinforces incidental correlations; neural prior provides medium acceleration but is bounded by the fidelity of the Taylor-expansion similarity metric. Our edge-routed contribution prior achieves the best performance on both metrics, confirming that responsibility-aware, contribution-weighted structural experience yields a substantially more efficient search prior than undifferentiated trajectory feedback or externally trained heuristics.

The reward distributions in Fig.~\ref{fig:rewards} reveal how the gain is obtained. Vanilla MCTS concentrates on low- and medium-reward candidates. The edge-routed contribution prior shifts probability mass toward higher-quality expressions, producing a pronounced secondary peak in the high-reward regime. Structural distillation therefore changes which regions of the program space are explored, rather than merely re-ranking an unchanged candidate set.

\subsection{PDE Recovery, Robustness, and General Symbolic Recovery}

Tab.~\ref{tab:mse} reports the dense-grid MSE across six PDE benchmarks. \method achieves the best performance on five of six problems, ranking second only to RSSP on Poisson3D. The largest gains occur on Wave2D and Wave3D, where solutions contain challenging compositions of exponential, trigonometric, and polynomial factors. On elliptic Poisson benchmarks, \method remains competitive or achieves the lowest error, confirming its effectiveness for both stationary and time-dependent PDEs. The smaller absolute differences on Advection and Diffusion are consistent with their simpler structures, although \method still obtains the best average MSE.

We next examine robustness under noisy and scarce observations. Gaussian noise $\alpha\in\{0\%,1\%,5\%,10\%\}$ is combined with retained-data ratios $r_d\in\{100\%,80\%,60\%,40\%\}$. A run is considered successful when the recovered expression is symbolically equivalent to the ground truth or achieves sufficiently low dense-grid error and PDE residual. Fig.~\ref{fig:robust} reports recovery rates over 100 trials per setting. Responsible-edge routing consistently improves recovery across all conditions, with the most pronounced gains under high noise and low data, indicating that counterfactual subtree attribution alleviates credit diffusion.

To evaluate generalization beyond PDE-constrained problems, we test \method on Nguyen, NguyenC, Jin, and Livermore symbolic regression suites, which cover polynomial, trigonometric, exponential, logarithmic, and nested nonlinear expressions. Fig.~\ref{fig:gt-sr} compares exact-recovery rates with MCTS-SR, RSSP, DSR, NGGP, and PySR. \method achieves the highest recovery on Nguyen, perfect recovery on NguyenC and Jin, and the best result on the more challenging Livermore suite. Together, Tab.~\ref{tab:mse} and Figs.~\ref{fig:robust}--\ref{fig:gt-sr} demonstrate that responsible-edge structural experience improves PDE discovery, robustness, and general symbolic regression performance.

\subsection{Coupled Multiphysics Systems}

We further evaluate SED-MCTS on nine coupled multiphysics problems, covering diffusion--convection, Joule-heating, and coupled parabolic systems. For each problem, the two unknown fields $f_1$ and $f_2$ are evaluated jointly under the coupled PDE constraints.
As shown in Table~\ref{compare}, SED-MCTS achieves the best results for both unknown fields on Joule Heating 1, 3, and 4, as well as Parabolic 3. It also obtains the lowest error for $f_2$ on Diffusion--convection and Parabolic 1, and for $f_1$ on Joule Heating 2. The improvements are especially pronounced for the thermal field in the strongly coupled Joule-heating cases. Although other methods perform better on several individual cases, particularly Joule Heating 5 and Parabolic 2, SED-MCTS demonstrates consistently competitive performance across different coupling structures. These results indicate that responsibility-aware structural experience can effectively guide symbolic search in coupled PDE systems.

\section{Conclusion}

This work presented SED-MCTS, an interpretable framework that converts terminal evaluations into reusable structural experience for symbolic PDE solution discovery. Context-preserving subtree neutralization identifies useful components, whose evidence is routed to responsible grammar edges and reused through a statistical prior and structural archive. Experiments demonstrate improved recovery accuracy, search efficiency, and robustness under noisy and scarce observations. Results on coupled multiphysics systems further confirm its ability to capture within-field and cross-field structural effects. Future work will reduce intervention costs and extend the framework to larger, more strongly coupled PDE systems.


\end{document}